\documentclass{article}

\usepackage{microtype}
\usepackage{graphicx}
\usepackage{subfigure}
\usepackage{booktabs} 

\usepackage{hyperref}
\usepackage{amsmath}
\usepackage{amssymb} 
\usepackage{bm}

\usepackage[dynnworkshop]{icml2022}

\renewcommand\url 

\icmltitlerunning{Parameter efficient dendritic-tree neurons outperform perceptrons}

\begin{document}

\twocolumn[
\icmltitle{Parameter efficient dendritic-tree neurons outperform perceptrons}



\icmlsetsymbol{equal}{*}

\begin{icmlauthorlist}
\icmlauthor{Ziwen Han}{equal,toronto}
\icmlauthor{Evgeniya Gorobets}{equal,toronto}
\icmlauthor{Pan Chen}{toronto}

\end{icmlauthorlist}

\icmlaffiliation{toronto}{Department of Computer Science, University of Toronto, Toronto, Canada}

\icmlcorrespondingauthor{Ziwen Han}{ziwen.han@mail.utoronto.ca}

\icmlkeywords{Machine Learning, ICML}

\vskip 0.3in
]



\printAffiliationsAndNotice{\icmlEqualContribution} 

\begin{abstract}
Biological neurons are more powerful than artificial perceptrons, in part due to complex dendritic input computations. Inspired to empower the perceptron with biologically inspired features, we explore the effect of adding and tuning input branching factors along with input dropout. This allows for parameter efficient non-linear input architectures to be discovered and benchmarked. Furthermore, we present a PyTorch module to replace multi-layer perceptron layers in existing architectures. Our initial experiments on MNIST classification demonstrate the accuracy and generalization improvement of dendritic neurons compared to existing perceptron architectures.
\end{abstract}

\section{Introduction}
Many artificial neural networks (ANNs) include variants of the perceptron [\citenum{rosenblatt1958perceptron}], which takes a linear combination of input signals and applies a nonlinear activation function to produce an output signal.
More recent neuroscience research has revealed that the dendrites of a biological neuron perform multiple complex nonlinear computations on their input signals [\citenum{london2005dendritic}], as opposed to a linear function. Furthermore, neuroscientists have advocated for incorporating dendritic features to improve existing ANN performance [\citenum{chavlis2021drawing}].
Empirically, Jones and Kording have demonstrated that a single dendritic neuron model with input repetition (k-trees) can reach accuracy similar to multi-layer perceptrons (MLPs) of similar parameter size on binary image classification tasks. [\citenum{jones2021might}]. Accordingly, we hypothesize multiple artificial dendritic neurons working in conjunction could be more powerful than their MLP counterpart beyond binary tasks. Current dendritic models are either limited by structural rigidity or fail to incorporate the tree-like structure of biological dendrites, which may not capture the full breadth of dendritic computation.

\section{Related Works}

\subsection{Neuron Models}
Multiple works have proposed ANNs based on neuron models that simulate dendritic input. These alternatives include dendrite morphological neural networks (DMNNs) [\citenum{ritter2003DMNN}], dendritic neural networks (DENNs) [\citenum{wu2018DENN}], the single dendritic neuron model (DNM) [\citenum{todo2014DNM}], and most recently the model by Jones and Kording based on a balanced tree structure [\citenum{jones2021might}].
Our model expands on the work of Jones and Kording by: (1) exploring the effect of generalizing the dendritic tree structure to allow tunable branching, dropout, and activations without k-tree redundancy; (2) using layers of dendritic neurons for non-binary classification tasks and evaluate overfitting; (3) attaching the layer of neurons to a CNN to evaluate performance as a perceptron replacement.

\subsection{Multi-Neurons}
Other studies have connected dendritic neurons in MLP-like architectures [\citenum{ritter2003DMNN}, \citenum{wu2018DENN}], including a hybrid DNM-CNN adaptation of the DNM model [\citenum{wang2022dendritic}]. Each of these multi-neuron architectures uses a fundamentally different neuron model. DNM neurons connect each dendritic branch to each input and rely on logical operations, while DMNNs utilize a different underlying mathematical structure. By contrast, our model enforces sparse, localized connections between dendrites and inputs in a tree structure, which more closely models the spatially-limited connections between biological neurons. The DENN model enforces dendrite-input sparsity, but uses one-layer dendrite trees [\citenum{wu2018DENN}]. Our dendritic trees are deeper, to more closely replicate the anatomy and complexity of biological neural networks.

\section{Methods}

\subsection{Model Architectures}
\begin{figure}[H]
 \centering
  \includegraphics[scale=0.35]{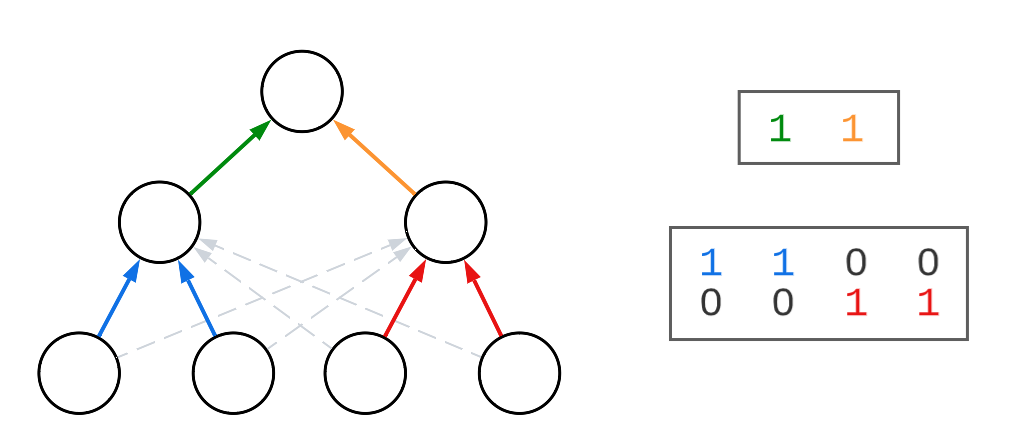}
  \caption{An example of the mask used to enforce the tree structure in our implementation. The tree structure on the left represents the dendritic neuron ($d=2, b=2$), rectangles on the right represent the mask matrices for each tree layer. The mask and weight matrices are multiplied element-wise. Colored arrows indicate preserved weights/connections, masked connections are represented by dashed gray arrows corresponding to zeroes. }
  \label{mask}
\end{figure}

Using PyTorch (1.10.0+cu111) [\citenum{pytorch}], we implemented a dendritic tree neuron based on the Jones-Kording single neuron model [\citenum{jones2021might}] (Figure \ref{diagram} in the Appendix). Our implementation generalizes the original architecture by allowing users to specify the branching factor and number of neurons in the \texttt{DendriticLayer}. Each \texttt{DendriticLayer} has a constant branching factor, but multiple instances of the module can be stacked together to achieve different branching per layer. To vectorize the \texttt{DendriticLayer}, we treat the inputs to all the neurons as a single input tensor, and we compute the next layer of all dendrite trees simultaneously. The tree structure is enforced by constructing a mask for the weight matrix at each layer (Figure \ref{mask}). Table \ref{architecture} describes the full architecture of the \texttt{DendriticLayer} module.

\begin{table*}[t]
 \centering
 \begin{tabular}{|c|l|} 
 \hline
 $\mathbf{y} = f^{(d)} \circ f^{(d-1)} \circ \cdots \circ f^{(1)} (\mathbf{x})$ & $\mathbf{x} \in \mathbb{R}^I, \mathbf{y} \in \mathbb{R}^O -$ the input and output tensors \\
    {$f^{(i)} (\mathbf{z}) = \text{LeakyReLU}((\mathbf{W}^{(i)} * \mathbf{M}^{(i)}) \mathbf{z} + \mathbf{b})$} & $f^{(i)}: \mathbb{R}^{I \cdot b^{-i+1}} \rightarrow \mathbb{R}^{I \cdot b^{-i}} -$ the function for the $i^{th}$ layer of the dendrite tree \\
 {} & $\mathbf{W}^{(i)}, \mathbf{M}^{(i)} \in \mathbb{R}^{I\cdot b^{-i} \times I \cdot b^{-i+1}} - $ the weight and mask matrices \\
 {} & $\mathbf{b}^{(i)} \in \mathbb{R}^{I\cdot b^{-i}} - $ the bias tensor  \\
 $I = Ob^d$ & $b -$ the branching factor; $d -$ the depth of the dendrite tree (\# edges) \\
 \hline
\end{tabular}
\caption{Equations describing the \texttt{DendriticLayer} architecture. The $*$ denotes element-wise matrix multiplication. The LeakyReLU activations used a negative slope value of $0.1$.}
\label{architecture}
\end{table*}

We used Jones and Kording's modified density gain Kaiming He initialization scheme to account for tree sparsity and stabilize training [\citenum{jones2021might}, \citenum{he2015}]. Each weight in the weight matrices is initially sampled: $\mathbf{W}^{(i)}_{jk} \overset{iid}{\sim} \mathcal{N}(0, \frac{2}{I\cdot b^{-i+1} \cdot \text{density}})$, where $\text{density} = \frac{I\cdot b^{-i+1}}{I\cdot b^{-i+1} \cdot I\cdot b^{-i}}$, and $I\cdot b^{-i+1}, I\cdot b^{-i}$ are the number of input and output units in the $i^{th}$ layer of the dendritic tree. 

Using the \texttt{DendriticLayer} as our base, we built several multi-layer neuron (MLN) architectures on top of it. (1) An \texttt{MLNBinaryClassifier}, which is a single neuron that predicts a binary output activated using a logistic sigmoid function. (2) An \texttt{MLNClassifier}, which repeats the same set of inputs to a specified number of neurons (one for each class) and returns the output of each neuron activated through a softmax function. (3) A \texttt{ConvMLN}, which runs the inputs through a CNN (described in Section 9.3 in the Appendix), then flattens the output of the convolutional layers and feeds it to a single layer of neurons for classification. All models are equipped with a tunable input dropout layer to be robust against overfitting [\citenum{2014dropout}]. 

\subsection{Computational Tasks and Controls}
We tested our single neuron models on binary classification tasks, using a subset of MNIST that consisted only of images labeled as "4" or "9" (referred to as 4-9 MNIST) [\citenum{deng2012mnist}]. We used the full MNIST dataset to test the classification capabilities of our multi-neuron architectures [\citenum{deng2012mnist}].

For controls, we constructed multi-layer perceptrons (MLPs) that performed the same tasks as each of our dendritic models. The architecture of the MLPs is described in Table \ref{mlp-architecture}. The number of hidden units ($h$) was modified in order to match the number of parameters in the dendritic models. The number of parameters in each model is listed in Tables \ref{binary-model-parameters}, \ref{multi-model-parameters}, and \ref{cnn-model-parameters} in the Appendix. Each model was initialized using the Kaiming He method, but without 
density gain since MLPs are not sparse [\citenum{he2015}].

\begin{table*}[t]
 \centering
 \begin{tabular}{|l|c|c|} 
 \hline
 {} & \textbf{Binary MLP} & \textbf{Multiclass MLP} \\
 \hline
 Equation & $y = \sigma\big(\mathbf{W}^{(2)}\big(\text{ReLU}(\mathbf{W}^{(1)} \mathbf{x} + \mathbf{b}^{(1)})\big) + b^{(2)})$ & $\mathbf{y} = \text{softmax}\big(\mathbf{W}^{(2)}\big(\text{ReLU}(\mathbf{W}^{(1)} \mathbf{x} + \mathbf{b}^{(1)})\big) + \mathbf{b}^{(2)})$ \\
 First Layer & $\mathbf{W}^{(1)} \in \mathbb{R}^{h \times bn}, \mathbf{b}^{(1)} \in \mathbb{R}^h$ & $\mathbf{W}^{(1)} \in \mathbb{R}^{h \times bn}, \mathbf{b}^{(1)} \in \mathbb{R}^h$ \\
 Second Layer &  $\mathbf{W}^{(2)} \in \mathbb{R}^{1 \times h}, b^{(2)} \in \mathbb{R}$ & $\mathbf{W}^{(2)} \in \mathbb{R}^{10 \times h}, \mathbf{b}^{(2)} \in \mathbb{R}^{10}$ \\
 Output & $y \in [0, 1]$ & $\mathbf{y} \in [0,1]^{10}$ \\
 \hline
\end{tabular}
\caption{Equations describing control MLP architectures}
\label{mlp-architecture}
\end{table*}

\subsection{Data Preprocessing and Results Analysis}
The standard MNIST dataset images are $28 \times 28$. Nearest neighbour upsampling was used to scale inputs to $32 \times 32$, to better fit the dendritic branching factors. For non-CNN architectures, input flattening was applied to create a $1024$-dimension tensor.

To aggregate results from multiple trials, the epoch with the lowest validation loss was taken from each trial as the best performance to evaluate at. Post-experiment statistical analysis was conducted using R [\citenum{R}].

\subsection{Model Training}
All models were trained on the Google Colab environment with CUDA for 100 epochs, using a batch size of 128. Every model was re-initialized and trained 10 times. The learning rates used for dendritic MLN and MLP models were 0.05 and 0.001, respectively. All models used either binary or categorical cross entropy loss. All models were trained using the Adam Optimizer [\citenum{kingma2014}]. 

\section{Results}

\subsection{Single Neuron Binary Classification}
We modified the Jones-Kording single neuron model to investigate the effects of branching factors ($b$) and dropout probabilities ($p$) (Table \ref{single-neuron-results}). The control MLPs with similar numbers of parameters are in Table \ref{mlp-binary-results}. Dropout models were tested with $p$ = 0.1, 0.2, 0.3, 0.4, 0.5, 0.6, but only the best-performing set of models are reported in each experiment.

\begin{table}
 \centering
 \begin{tabular}{|c|c|c|c|c|c|c|} 
 \hline
 {\textbf{Single Neuron}} & $b$ & $p$ & {\textbf{Train Acc.}} & {\textbf{Val. Acc.}} \\ 
 \hline
 1 & 2 & 0 & 0.91 $\pm$ 0.06 & 0.85 $\pm$ 0.09 \\ 
 2 & 4 & 0 & 0.94 $\pm$ 0.06 & 0.89 $\pm$ 0.06 \\ 
 3 & 32 & 0 & 0.99 $\pm$ 0.02 & 0.91 $\pm$ 0.02 \\ 
 4 & 2 & 0.4 & 0.85 $\pm$ 0.07 & 0.89 $\pm$ 0.07 \\ 
 5 & 4 & 0.5 & 0.89 $\pm$ 0.03 & 0.92 $\pm$ 0.02 \\ 
 6 & 32 & 0.3 & 0.96 $\pm$ 0.02 & 0.92 $\pm$ 0.02 \\
 \hline
\end{tabular}
\caption{Mean performance $\pm$ standard deviation of a single dendritic neuron (\texttt{MLNBinaryClassifier}) on 4-9 MNIST. Models differ in their branching factor ($b$) and dropout probability ($p$).}
\label{single-neuron-results}
\end{table}

\begin{table}
 \centering
 \begin{tabular}{|c| c| c| c| c| c| c|} 
 \hline
 \textbf{Binary MLP} & $h$ & $p$ & \textbf{Train Acc.} & \textbf{Val. Acc.} \\ 
 \hline
 1 & 3 & 0 & 0.95 $\pm$ 0.15 & 0.88 $\pm$ 0.13 \\ 
 2 & 2 & 0 & 0.86 $\pm$ 0.22 & 0.79 $\pm$ 0.19 \\
 3 & 3 & 0.6 & 0.95 $\pm$ 0.02 & 0.92 $\pm$ 0.01 \\
 4 & 2 & 0.4 & 0.97 $\pm$ 0.01 & 0.91 $\pm$ 0.01 \\
 \hline
\end{tabular} \\
\caption{Mean performance $\pm$ standard deviation of binary MLPs on 4-9 MNIST. $h =$ number of hidden units; $p = $ dropout probability. MLPs with $h=3$ were controls for neurons with $b=2$; MLPs with $h=2$ were controls for neurons with $b=4, 32$ (smallest possible two-layer MLP).}
\label{mlp-binary-results}
\end{table}

\subsection{Multi-Neuron Classification}
To test non-binary classification, we integrated multiple neurons in a layer (\texttt{MLNClassifier}). Each neuron connected to the same set of inputs and was expected to predict a single MNIST digit. The experimental setup identical to the binary case, as described in Sections 3.4, 4.1. The average performance of each multi-neuron model and each control MLP is reported in Tables \ref{multi-neuron-results} and \ref{mlp-multi-results}.

\label{multi-neuron-results}
\begin{table}
 \centering
 \begin{tabular}{|c|c|c|c|c|c|c|} 
 \hline
 {\textbf{MLN}} & $b$ & $p$ & {\textbf{Train Acc.}} & {\textbf{Val. Acc.}} \\ 
 \hline
 1 & 2 & 0 & 0.75 $\pm$ 0.07 & 0.59 $\pm$ 0.04 \\ 
 2 & 4 & 0 & 0.90 $\pm$ 0.06 & 0.67 $\pm$ 0.05 \\ 
 3 & 32 & 0 & 0.94 $\pm$ 0.05 & 0.77 $\pm$ 0.02 \\ 
 4 & 2 & 0.2 & 0.77 $\pm$ 0.06 & 0.62 $\pm$ 0.04 \\ 
 5 & 4 & 0.6 & 0.79 $\pm$ 0.07 & 0.75 $\pm$ 0.06 \\ 
 6 & 32 & 0.4 & 0.97 $\pm$ 0.02 & 0.82 $\pm$ 0.02 \\
 \hline
\end{tabular}
\caption{Mean performance $\pm$ standard deviation of a ten-neuron single-layer model on MNIST. The models differ in their branching factor ($b$) and dropout probability ($p$).}
\label{multi-neuron-results}
\end{table}

\begin{table}
 \centering
 \begin{tabular}{|c| c| c| c| c| c| c|} 
 \hline
 \textbf{Multiclass MLP} & $h$ & $p$ & \textbf{Train Acc.} & \textbf{Val. Acc.} \\ 
 \hline
 1 & 30 & 0 & 0.93 $\pm$ 0.07 & 0.63 $\pm$ 0.07 \\
 2 & 14 & 0 & 0.87 $\pm$ 0.08 & 0.58 $\pm$ 0.07 \\
 3 & 11 & 0 & 0.87 $\pm$ 0.11 & 0.58 $\pm$ 0.04\\
 4 & 30 & 0.5 & 0.92 $\pm$ 0.04 & 0.66 $\pm$ 0.04 \\
 5 & 14 & 0.1 & 0.89 $\pm$ 0.07 & 0.61 $\pm$ 0.06 \\
 6 & 11 & 0.3 & 0.86 $\pm$ 0.08 & 0.59 $\pm$ 0.07 \\
 \hline
\end{tabular} \\
\caption{Mean performance $\pm$ standard deviation of MLPs on MNIST. $h =$ number of hidden units; $p = $ dropout probability. MLPs with $h=30, 14, 11$ were controls for MLNs with $b=2, 4, 32$, respectively.}
\label{mlp-multi-results}
\end{table}

\subsection{Dendritic Models with CNNs}
To explore the potential as a tunable, modular MLP replacement in existing architectures, we connected 10 neurons to a simple CNN architecture for MNIST classification (\texttt{ConvMLN}) as a replacement to the usual perceptrons. This experimental setup is described in Sections 3.4, 4.1. The performance of each CNN-MLN and control CNN-MLP is listed in Tables \ref{cnn-mln-results}and \ref{cnn-mlp-results}.

\begin{table}
 \centering
 \begin{tabular}{|c|c|c|c|c|c|c|} 
 \hline
 {\textbf{CNN-MLN}} & $b$ & $p$ & {\textbf{Train Acc.}} & {\textbf{Val. Acc.}} \\ 
 \hline
 1 & 2 & 0 & 0.82 $\pm$ 0.07 & 0.61 $\pm$ 0.08 \\ 
 2 & 4 & 0 & 0.96 $\pm$ 0.03 & 0.74 $\pm$ 0.04 \\ 
 3 & 16 & 0 & 0.99 $\pm$ 0.02 & 0.78 $\pm$ 0.03 \\ 
 4 & 2 & 0.1 & 0.77 $\pm$ 0.07 & 0.64 $\pm$ 0.07 \\ 
 5 & 4 & 0.2 & 0.93 $\pm$ 0.04 & 0.81 $\pm$ 0.04 \\ 
 6 & 16 & 0.5 & 0.95 $\pm$ 0.02 & 0.83 $\pm$ 0.03 \\
 \hline
\end{tabular}
\caption{Mean performance $\pm$ standard deviation of CNN-MLN models on MNIST classification. The models differ in their branching factor ($b$) and dropout probability ($p$).}
\label{cnn-mln-results}
\end{table}

\begin{table}
 \centering
 \begin{tabular}{|c|c|c|c|c|c|c|} 
 \hline
 {\textbf{CNN-MLP}} & $h$ & $p$ & {\textbf{Train Acc.}} & {\textbf{Val. Acc.}} \\ 
 \hline
 1 & 29 & 0 & 0.97 $\pm$ 0.04 & 0.7 $\pm$ 0.05 \\ 
 2 & 16 & 0 & 0.99 $\pm$ 0.03 & 0.7 $\pm$ 0.06 \\ 
 3 & 11 & 0 & 0.94 $\pm$ 0.06 & 0.64 $\pm$ 0.02 \\ 
 4 & 29 & 0.1 & 0.99 $\pm$ 0.02 & 0.74 $\pm$ 0.06 \\ 
 5 & 16 & 0.4 & 0.98 $\pm$ 0.02 & 0.74 $\pm$ 0.06  \\ 
 6 & 11 & 0.1 & 0.96 $\pm$ 0.06 & 0.67 $\pm$ 0.07 \\
 \hline
\end{tabular}
\caption{Mean performance $\pm$ standard deviation of CNN-MLPs on MNIST. $h =$ \# of hidden units; $p = $ dropout probability. CNN-MLPs with $h=29, 16, 11$ were controls for CNN-MLNs with $b=2, 4, 16$, respectively.}
\label{cnn-mlp-results}
\end{table}

\subsection{Results Summary}
Both increasing branching factor and introducing dropout significantly improved performance. The best performing dendritic models were two layers deep, with $b = \sqrt{i}$, where $i$ is number of inputs to each neuron and $p \geq 0.1$. 
We hypothesize deeper trees led to vanishing gradients; shallow dendritic trees trained and performed better despite having fewer parameters.
In all experiments, dendritic neuron models with $b=2$ perform worse than their control MLPs, but neurons with $b > 2$ surpass their control MLPs in terms of validation performance and robustness to overfitting, both with and without dropout. This shows dendritic trees can improve performance while maintaining parameter efficiency relative to MLP counterparts.

\section{Discussion}
We created an encapsulated generalized dendritic-tree neuron inspired by Jones and Kording, then combined multiple of them analagous to multi-layer perceptrons, evaluated on the MNIST dataset [\citenum{deng2012mnist}, \citenum{jones2021might}]. We investigated the effect of adding input dropout and increasing branching factor (decreasing dendritic tree depth) and found both to have a positive effect on performance. Furthermore, we evaluated our model performance relative to MLPs of similar parameter size on binary MNIST 4-9 classification and on full MNIST multi-class classification, both as direct input and when attached to a simple convolutional neural network. Even without using the k-tree repeated attachment in Jones-Kording, our model is able to outperform parameter matched MLPs. Though all models overfit on the training data, the sparse tree structure reduces it relative to an MLP.

Our work demonstrates the potential for a dendritic tree neuron of similar parameter size to replace an MLP in a feed-forward layer, as it gives better robustness to overfitting, better performance, and potentially better interpretability due to the hierarchical tree structure. This work also exhibits the power of input non-linearity over classical perceptron architecture.

\section{Limitations and Future Work}
Computational optimization: The current implementation in PyTorch [\citenum{pytorch}] utilizes masks on full weight matrices, which add unnecessary computation to a sparse structure. When the dendritic tree structure is deep, we cannot efficiently parallelize operations relative to parameter-matched shallower MLPs as the signal is depth-wise sequentially passed through the tree. However, the theoretical reduction in the number of operations and parameters still holds. For applicability, future work should focus on optimizing the practical computations by taking advantage of the tree structure.

Model Tuning: Our current experiments involve naively attaching the dendritic tree structure to a flattened input. Up/down-sampling is required for tree-structured vectorized computations in each layer to match the branching factor, adding rigidity. Methods for attaching tree-dendrites to input vectors to take advantage of locality structures are yet to be investigated. Furthermore, exploration can be made to automatically tune the branching factor or explore non-balanced branching as a dynamic feature.

General Applications: It remains to be discovered how well the dendritic module performs as a perceptron replacement on tasks outside of MNIST classification. Our experiments focused mainly on shallow networks; future experiments can investigate if a similar trend to current results is observed on larger, deeper, state of the art models. We also focused on testing parameter efficiency with multi-layer perceptrons against the dendritic neuron counterpart, which may result in subpar optimal accuracy. Further exploration is required to evaluate how extreme we can reduce the number of parameters in the input tree structure relative to a perceptron while preserving performance.

\section{Conclusion}
To augment artificial neural networks with biologically inspired input non-linearity, we created a general dendritic neuron module for tunable branching and dropout. Compared against multi-layer perceptrons, our implementation maintains theoretical parameter efficiency with better performance on tuned branching. Promising evaluation on MNIST classification independently and modularly highlights the possibility of optimizing and incorporating these structures into existing architectures for better accuracy and generalization. These results suggest input non-linearity structure as a possible direction to explore for dynamic neural networks.

\section{Acknowledgements}
We would like to thank Harris Chan from the University of Toronto for his encouragement and support on this project.

\newpage
\bibliography{main}
\bibliographystyle{icml2021}

\newpage
\section{Appendix}


\subsection{Reproducibility}
PyTorch implementation and experiment data linked here: 
github.com/zw123han/DendriticNeuralNetwork

\subsection{Model Diagrams}
\begin{figure}[H]
 \centering
  \includegraphics[scale=0.65]{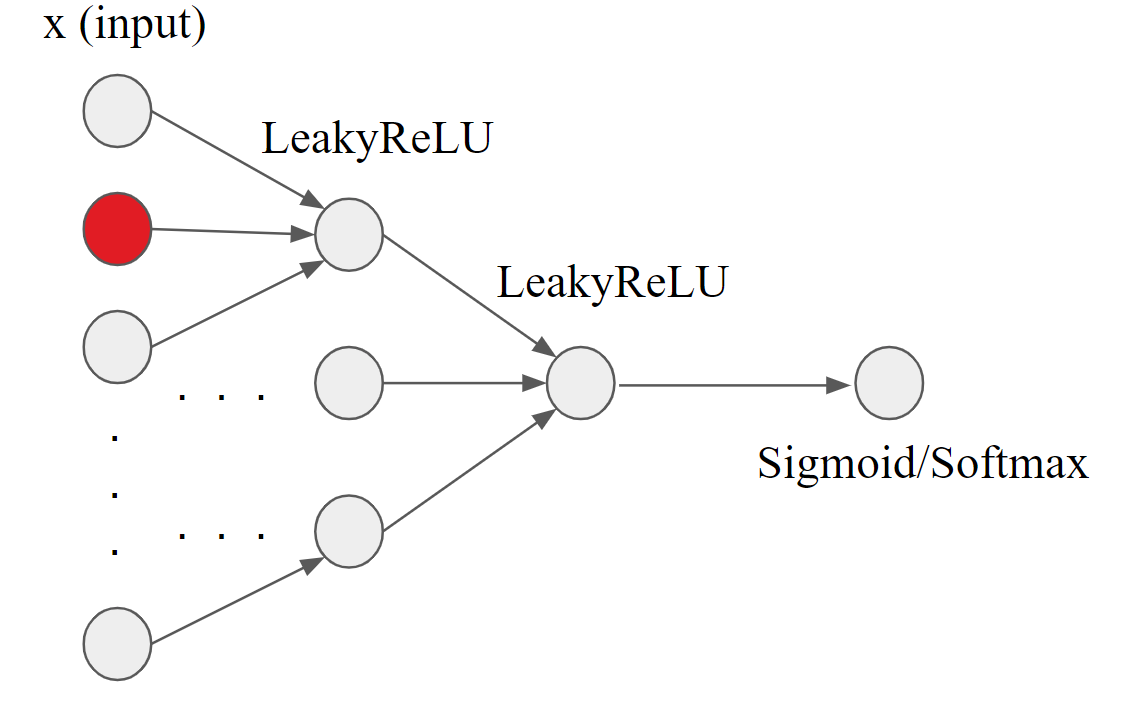}
  \caption{Architecture of a single dendritic neuron with branching $b=3$ and flattened input size $3^k$. Red represents an input dropout, which is stochastically masked at training time using the PyTorch dropout layer implementation. A LeakyReLU is applied to each layer as activation to model the input non-linearity of dendrites. The output layer, if appropriate, uses a sigmoid/softmax activation to create a probability vector.}
  \label{diagram}
\end{figure}


\subsection{CNN Architecture}

An identical and simple CNN was attached to all classifiers. The CNN consisted of three sets of convolutions. Each convolutional layer had kernel size of 5 and a padding of 2. The number of filters in the first layer was 4, the number of filters in the second layer was 8, the number of filters in the third layer was 16. Each convolutional layer was followed by a MaxPool layer (kernel size = 2, stride = 2), a BatchNorm layer, and a ReLU activation layer. 

The initial input to the CNN was a $B \times 1 \times 32 \times 32$ tensor, where $B$ was the batch size, $1$ is the number of input channels (grayscale), and $32 \times 32$ are the dimensions of the upsampled MNIST digits. Each set of convolutions and pooling cut the size of the image in half ($32 \rightarrow 16 \rightarrow 8 \rightarrow 4$). The final output of the CNN was 16 4$\times$4 maps, which were flattened into a single 256-unit input before it was given to the dendritic classifier.

\subsection{Model Parameters}

\begin{table}
 
 \centering
 \begin{tabular}{|c| c| c| c|} 
 \hline
 \textbf{Model} & \textbf{Weights} & \textbf{Biases} & \textbf{Total} \\ 
 \hline
 Neuron Models 1, 4 ($b=2$) & 2046 & 1023 & 3069 \\ 
 MLPs 1, 3 ($h=3$) & 3075 & 4 & 3079 \\ 
 Neuron Models 2, 5 ($b=4$) & 1364 & 85 & 1449 \\ 
 Neuron Models 3, 6 ($b=32$) & 1056 & 33 & 1089 \\ 
 MLPs 2, 4 ($h=2$) & 2050 & 3 & 2053 \\ 
 \hline
\end{tabular} \\
\caption{Single Neuron Experiments. Parameter computations for single-neuron models and their MLP controls. All models ran on MNIST images upsampled to 32$\times$32 images with binary output. MLPs 1, 3 ($h=3$) have approximately the same number of parameters as Neuron Models 1, 4 ($b=2$), and thus serve as controls for these models. Similarly, MLPs 2, 4 ($h=2$) are the controls for Neuron Models 2, 3, 5, 6 ($b=4, 32$).}
\label{binary-model-parameters}
\end{table}

\begin{table}
 
 \centering
 \begin{tabular}{|c| c| c| c|} 
 \hline
 \textbf{Model} & \textbf{Weights} & \textbf{Biases} & \textbf{Total} \\ 
 \hline
 MLNs 1, 4 ($b=2$) & 20,460 & 10,230 & 30,690 \\ 
 MLPs 1, 4 ($h=30$) & 31,020 & 40 & 31,060 \\ 
 MLNs 2, 5 ($b=4$) & 13,640 & 850 & 14,490 \\ 
 MLPs 2, 5 ($h=14$) & 14,476 & 24 & 14,500 \\ 
 MLNs 3, 6 ($b=32$) & 10,560 & 330 & 10,890 \\ 
 MLPs 3, 6 ($h=11$) & 11,374 & 21 & 11,395 \\ 
 \hline
\end{tabular} \\
\caption{Multi-Neuron Single-Layer Experiments. Parameter computations for multi-neuron single-layer models (MLNs) and their MLP counterparts. All models ran on MNIST images upsampled to 32$\times$32 images, for a total 1024 inputs. Each model had 10 outputs. MLPs 1, 4 ($h=30$) have approximately the same number of parameters as MLNs 1, 4 ($b=2$), and thus serve as controls for these models. Similarly, MLPs 2, 5 ($h=14$) are the controls for MLNs 2, 5, ($b=4$), and MLPs 3, 6 ($h=11$) are the controls for MLNs 3, 6 ($b=32$).}
\label{multi-model-parameters}
\end{table}

\begin{table}
 
 \centering
 \begin{tabular}{|c| c| c| c|} 
 \hline
 \textbf{Model} & \textbf{Weights} & \textbf{Biases} & \textbf{Total} \\ 
 \hline
 CNN-MLNs 1, 4 ($b=2$) & 5100 & 2550 & 7650 \\ 
 CNN-MLPs 1, 4 ($h=29$) & 7714 & 39 & 7753 \\ 
 CNN-MLNs 2, 5 ($b=4$) & 3400 & 850 & 4250 \\ 
 CNN-MLPs 2, 5 ($h=16$) & 4256 & 26 & 4282 \\ 
 CNN-MLNs 3, 6 ($b=16$) & 2720 & 170 & 2890 \\ 
 CNN-MLPs 3, 6 ($h=11$) & 2926 & 21 & 2947 \\ 
 \hline
\end{tabular} \\
\caption{CNN Experiments. Parameter computations for CNN classification models. All models ran on MNIST images upsampled to 32$\times$32 images, for a total 1024 inputs.  Each classifier was attached to the same CNN (described in Section 7.4), which output a 256-dim tensor. Thus, each classifier had 256 inputs and 10 outputs. CNN-MLPs 1, 4 ($h=29$) have approximately the same number of parameters as CNN-MLNs 1, 4 ($b=2$), and thus serve as controls for these models. Similarly, CNN-MLPs 2, 5 ($h=16$) are the controls for CNN-MLNs 2, 5, ($b=4$), and CNN-MLPs 3, 6 ($h=11$) are the controls for CNN-MLNs 3, 6 ($b=16$).}
\label{cnn-model-parameters}
\end{table}

\end{document}